\documentclass{article} % For LaTeX2e
\usepackage{iclr2024_conference,times}

\makeatletter
\def\@fnsymbol#1{\ensuremath{\ifcase#1\or \dagger\or \ddagger\or
   \mathsection\or \mathparagraph\or \|\or **\or \dagger\dagger
   \or \ddagger\ddagger \else\@ctrerr\fi}}
\makeatother
    
\usepackage{amsmath,amsfonts,bm}

\def\eqref#1{equation~\ref{#1}}
\def\1{\bm{1}}

\def\vzero{{\bm{0}}}

\def\vx{{\bm{x}}}

\def\mI{{\bm{I}}}

\DeclareMathAlphabet{\mathsfit}{\encodingdefault}{\sfdefault}{m}{sl}
\SetMathAlphabet{\mathsfit}{bold}{\encodingdefault}{\sfdefault}{bx}{n}

\def\sD{{\mathbb{D}}}
\usepackage{hyperref}
\usepackage{url}
\usepackage{graphicx} 
\usepackage{booktabs}

\usepackage{listings}
\usepackage{xcolor} % for custom colors

\hypersetup{
    colorlinks=true,
    linkcolor=red,
    citecolor=cyan,
    filecolor=magenta,      
    urlcolor=cyan,
}

\usepackage{amsmath,amsfonts, mathtools, bm, bbm, nicefrac}
\newcommand{\ptheta}{p_{\theta}}

\newcommand{\mcal}[1]{\mathcal{#1}}

\def\vzero{{\bm{0}}}

\def\vx{{\bm{x}}}

\def\mI{{\bm{I}}}

\DeclareMathAlphabet{\mathsfit}{\encodingdefault}{\sfdefault}{m}{sl}
\SetMathAlphabet{\mathsfit}{bold}{\encodingdefault}{\sfdefault}{bx}{n}
\def\sD{{\mathbb{D}}}
\title{ChatAnything: \\
Facetime Chat with LLM-Enhanced Personas}

\author{Yilin Zhao$^{1}$\thanks{Equal contribution.} \quad
Xinbin Yuan$^{1}$\footnotemark[1] \quad
Shanghua Gao$^{1}$\footnotemark[1] \quad
\AND
\vspace{-25px}
\\
Zhijie Lin$^2$ \quad
Qibin Hou$^1$\thanks{Project lead.} \quad
Jiashi Feng$^{2}$~ \quad
Daquan Zhou$^{2}$\thanks{Corresponding to: \texttt{zhoudaquan21@gmail.com}.}~~\footnotemark[2]
\\
\\
$^1$Nankai University $^2$ByteDance  \\
}

\iclrfinalcopy % Uncomment for camera-ready version, but NOT for submission.
\begin{document}

\maketitle

% \footnote{* Equal contribution.}

\begin{abstract}
In this technical report, we target generating anthropomorphized personas for LLM-based characters in an online manner, including visual appearance, personality and tones, with only text descriptions. To achieve this, we first leverage the in-context learning capability of LLMs for personality generation by carefully designing a set of system prompts. We then propose two novel concepts: the mixture of voices (MoV) and the mixture of diffusers (MoD) for diverse voice and appearance generation. For MoV, we utilize the text-to-speech (TTS) algorithms with a variety of pre-defined tones and select the most matching one based on the user-provided text description automatically. 
For MoD, we combine the recent popular text-to-image generation techniques and talking head algorithms to streamline the process of generating talking objects. We termed the whole framework as \emph{ChatAnything}.
With it, users could be able to animate anything with any personas that are anthropomorphic using just a few text inputs. However, we have observed that the anthropomorphic objects produced by current  generative models are often undetectable by pre-trained face landmark detectors, leading to  failure of the face motion generation, even if these faces possess human-like appearances because those images are nearly seen during the training (e.g., OOD samples). To address this issue, we  incorporate pixel-level guidance to infuse human face landmarks during the image generation phase. To benchmark these metrics, we have built an evaluation dataset. Based on it, we verify that the detection rate of the face landmark is significantly increased from 57.0\% to 92.5\% thus allowing automatic face animation based on generated speech content. 
% In the whole process, only texts are needed for the definition of static images and the driving signal. 
The code and more results can be found at \url{https://chatanything.github.io/}\footnote{Work in progress}.
\end{abstract}

\section{Introduction}

\begin{figure}[!h]
\label{fig:driving_samples}
\begin{center}

\includegraphics[width=0.95\textwidth]{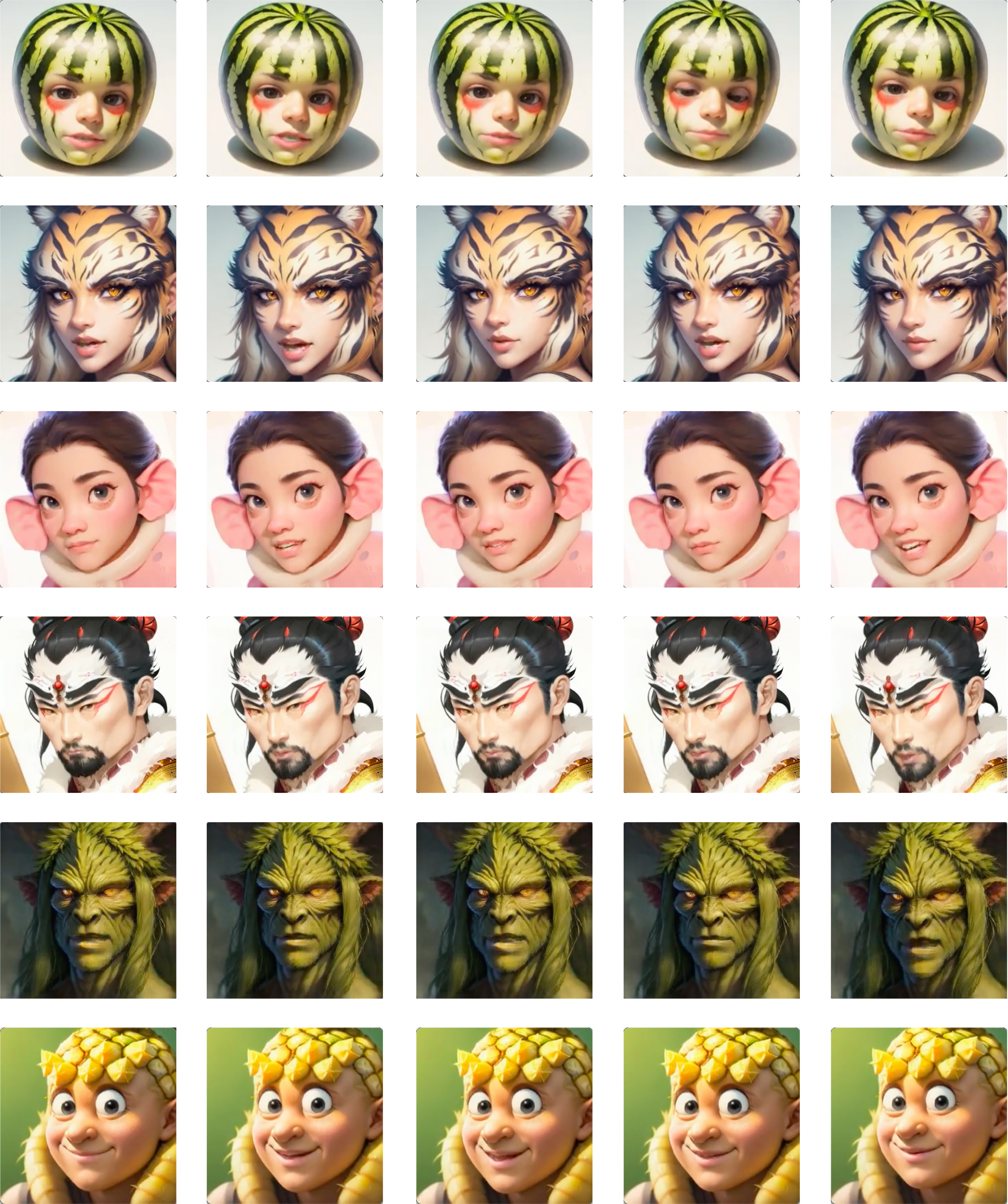}

\end{center}
\caption{Text `what' you want to chat with! A purely text-based LLM-enhanced Facetime framework that bridges the distribution gap between the pre-trained generative models and the pre-trained talking-head driving models. Face expression generation is based on the generated portraits with SadTalker \citep{zhang2023sadtalker}.}
\end{figure}

% \begin{figure}[!h]
% \label{fig:sample_generated_faces}
% \begin{center}

% \includegraphics[width=1\textwidth]{figures/teaser.pdf}

% \end{center}
% \end{figure}

Recent advancements in the field of large language models (LLMs) \citep{wei2022emergent,openai2023gpt4} have positioned them at the forefront of academic discussion due to their remarkable generalization and in-context learning capacities. Such models facilitate interactions on a plethora of topics, offering users an experience that closely mirrors human-like conversation.
In this report, we explore a new framework that could generate LLM-enhanced personas with customized personalities, voices, and visual appearances. 
For the personality injection, we alleviate the in-context learning capability of LLMs where we define a carefully designed system prompt to generate customized characters based on the user text description on the objects that they are expecting to chat with. 
For the voices, we first create a pool of voice modules based on the text-to-speech (TTS) APIs\footnote{https://learn.microsoft.com/en-us/azure/ai-services/speech-service/}. Then, we assign each tone a specific description such that the LLMs could select the most matching tones according to the user text inputs in a similar way as LangChian\citep{Chase_LangChain_2022} and Visual-ChatGPT \citep{wu2023visual}. We term this module as the mixture-of-voices (MoV).
The main challenge comes from the visual appearance of speech-driven talking motions and expressions.
We thus explore utilizing the recent popular talking head algorithms. Specifically, with input audio, talking portrait animation is defined by synthesizing a group of frames based on a single image to simulate the talking context that is synchronized with the audio signal \citep{chen2020talking,kr2019towards}. Despite its breakthroughs in linking audio information to expressive facial motions, the input image formats are not explored as much as those in motion synchronization. In the report, we explore utilizing recent popular generative models to further simplify the full pipeline such that only text inputs are needed. For a conventional talking head framework \citep{narayana4573122pixels}, two inputs are needed: 1. an audio clip and 2. an image for visual appearance rendering. Intuitively, both the two inputs can be text-driven only: the audio inputs can be generated based on text inputs with  MoV, and the image inputs can be text-driven by utilizing recent popular Text-To-Image (T2I) models \citep{rombach_high-resolution_2022,yang2023diffusion}.  However, we empirically found that the images generated with recent popular diffusion models cannot be used as the source image for popular talking head models. We then systematically examine this by generating 400 anthropomorphic samples from 8 categories including human-realistic, animals, fruits, plants, office, bags, clothes, and cartoon styles.  Some of the examples are shown in Fig. \ref{fig:sample_generated_faces}. Among all those samples, only 30\% of images can be detected by the recent state-of-the-art talking head model SadTalker \citep{zhang2023sadtalker} with RetinaFace landmark detector \citep{deng2019retinaface}. This can be interpreted in that the generated images are not seen during the training of the face detectors. The data distribution of the generated data is not aligned with the training data of the face synthesis module and is thus regarded as out-of-distribution data. For example, if we use a face detector pre-trained with anime data, the landmark detection rate will be increased. A direct way to improve this is to fine-tune the model with images used for the training of diffusion-based generative models \citep{rombach_high-resolution_2022}. However, this is computationally heavy, and many of the training pipelines of the state-of-the-art talking head algorithms are not publicly available \citep{zhang2023sadtalker,guan2023stylesync}. We are thus motivated to find a zero-shot method to bridge the distribution gap between the pre-trained generative models and the talking head models without reproducing the training procedure of those pre-trained models. 

\begin{figure}[t]
\label{fig:pipeline}
\begin{center}
\includegraphics[width=1.02\textwidth]{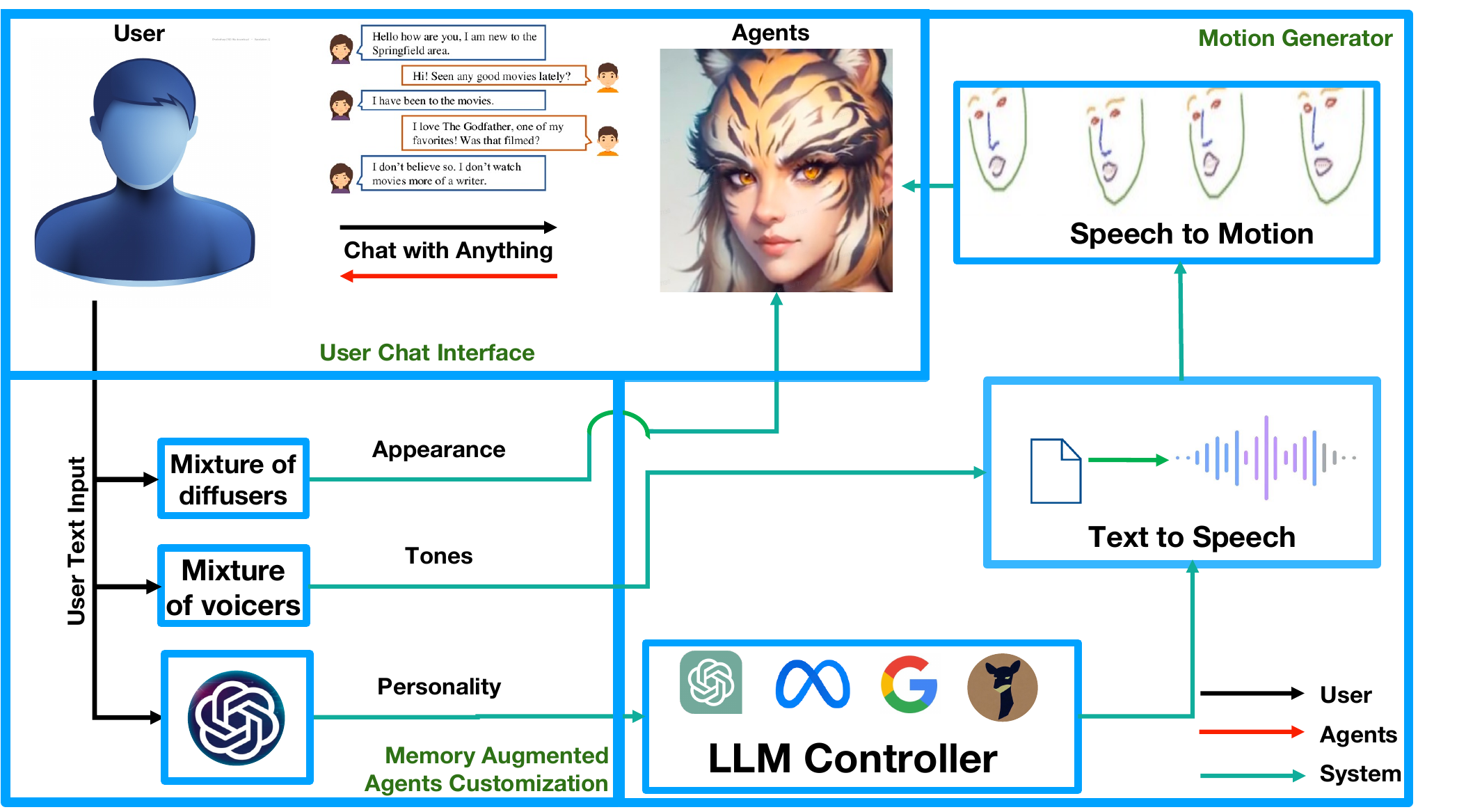}
\end{center}
\caption{The overall pipeline of ChatAnything. ChatAnything framework includes four key components: 1. a portrait generation component; 2. a personality generation component; 3, a voice generation component, and 4. a face-driving component. To further reinforce the customization freedom of the generated personas, we introduce two novel concepts: the mixture of diffusers (MoD) and the mixture of voices (MoV) where the style of the appearance and the tones can be customized based on the user text description. For the details of the design of each component, please refer to Sec. \ref{subsec:components}.}
\end{figure}

The diffusion process is an auto-regressive process where the noise is removed gradually conditioned on the feature maps from the previous timestep. Many image editing methods utilize this property to achieve photorealistic images with desirable edited properties \citep{chefer2023attend,zhou2023maskdiffusion,liew2022magicmix}. 
Motivated by this, 
we seek to find a pixel-level landmark injection method that could force the landmark trajectory to be detected without an obvious impact on visual appearance in a zero-shot manner. 
For meticulous injection of facial landmark trajectories, we harness the iterative denoising procedure intrinsic to pre-trained diffusion models, wherein the face landmark features are embedded in the starting noise before the denoising process. 
Nevertheless, our investigations elucidate a challenging equilibrium between landmark conservation and text concept infusion. Specifically, overemphasized landmark pixels often result in generated imagery that is incongruent with the textual descriptors, compromising aesthetic appeal. Conversely, an understated emphasis on landmark pixels yields overly abstract renditions, resulting in undetectable face landmarks with pre-trained face detectors.
In certain cases, there is no optimal sweet point or hard to search. Motivated by recent concept injection and erasing techniques \citep{zhang2023adding,zhang2023forget}, we seek to utilize cross-attention blocks to reinforce the overall structure information. 
The hybrid implementation of hard injection methodologies with structural concept injection gives the best trade-off between landmark injection and visual appearance. Similar to MoV, we also prepare a pool of image generation models based on latent diffusion model (LDM) \cite{rombach2022high} with different styles\footnote{The models are downloaded from \href{https://civitai.com/models/}{Civitai}}. During inference, the LLMs will select the most relevant one based on the text description and we term this as the mixture of diffusers (MoD). The key contributions of our work can be delineated as follows:

\begin{itemize}
    \item We introduce a novel framework dedicated to the generation of LLM-enhanced personas exclusively from textual inputs. Predicated on user-specified keywords, our method synthesizes both a portrait and an associated personality and voice, facilitating meaningful user interaction with the resultant persona.
    \item We introduce a zero-shot approach designed to harmonize the distribution between pre-trained generative models and per-trained talking head models. This alignment ensures the production of expressive facial movements based on the synthesized avatar portrait.
    \item We propose an evaluation dataset to quantify the alignment between the generative models and the talking-head models.
\end{itemize}

\begin{figure}[!h]
\label{fig:comparison}
\begin{center}
\includegraphics[width=1\textwidth]{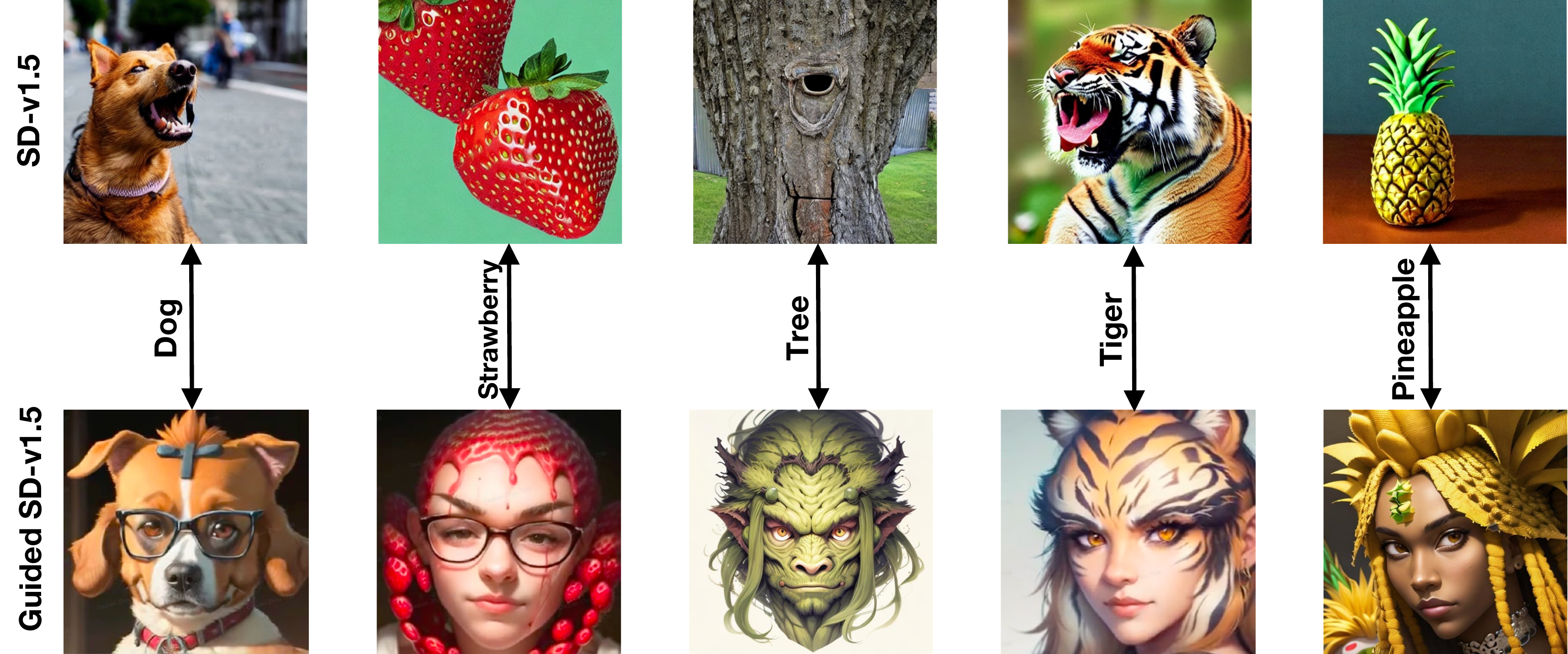}
\end{center}
\caption{Impact of landmark guidance during the diffusion process. As shown in the first row, directly apply SD1.5 for portrait generation tends to produce abstract face images. Those images are rarely seen during the training of the talking head models and thus cannot be used as the input for facial expression generation. Differently, after applying the proposed techniques in ChatAnything (including the face landmark guidance, prompt engineering, and LoRA fine-tuning for aesthetics improvements),  the model tends to generate more anthropomorphic images with high visual quality that can be used as the input for pre-trained talking head models. }
\end{figure}

\section{Method}

In this section, we  introduce  the details for the pipeline of the ChatAnything framework, which includes four main blocks:
\begin{enumerate}
\setlength\itemsep{0em}
  \item A LLM-based control module that initializes the personality of the text-described persona from the user. It is also used to manage the system operation and call applications based on the interactions with the users.
  \item A portrait initializer that generates the reference image for the persona. It includes a mixture of finetuned diffusion models (MoD) along with their LoRA module (if applicable). Each model is specialized in generating a specific style of images. The most matching model will be called automatically based on the user text description of the persona vis LLM.
  \item A mixture of text-to-speech modules (MoV) that converts the text input from the personas to speech signals with customized tones. The selection is done automatically based on the user text descriptions via LLM.
  \item A Motion generation module that takes in the speech signal and drives the generated image. 
\end{enumerate}

The overall system architecture design is shown in Fig. \ref{fig:pipeline}. 
In the rest of this section, we first give a preliminary on the working mechanism of diffusion models in Sec.\ref{subsec:components}. This foundation will serve as the premise for elucidating our motivations behind specific design choices. Subsequently, in Sec.\ref{subsec:components}, we will provide an exhaustive exposition of the design intricacies of each constituent component.

\begin{figure}[t]
\label{fig:mod}
\begin{center}
\includegraphics[width=0.8\textwidth]{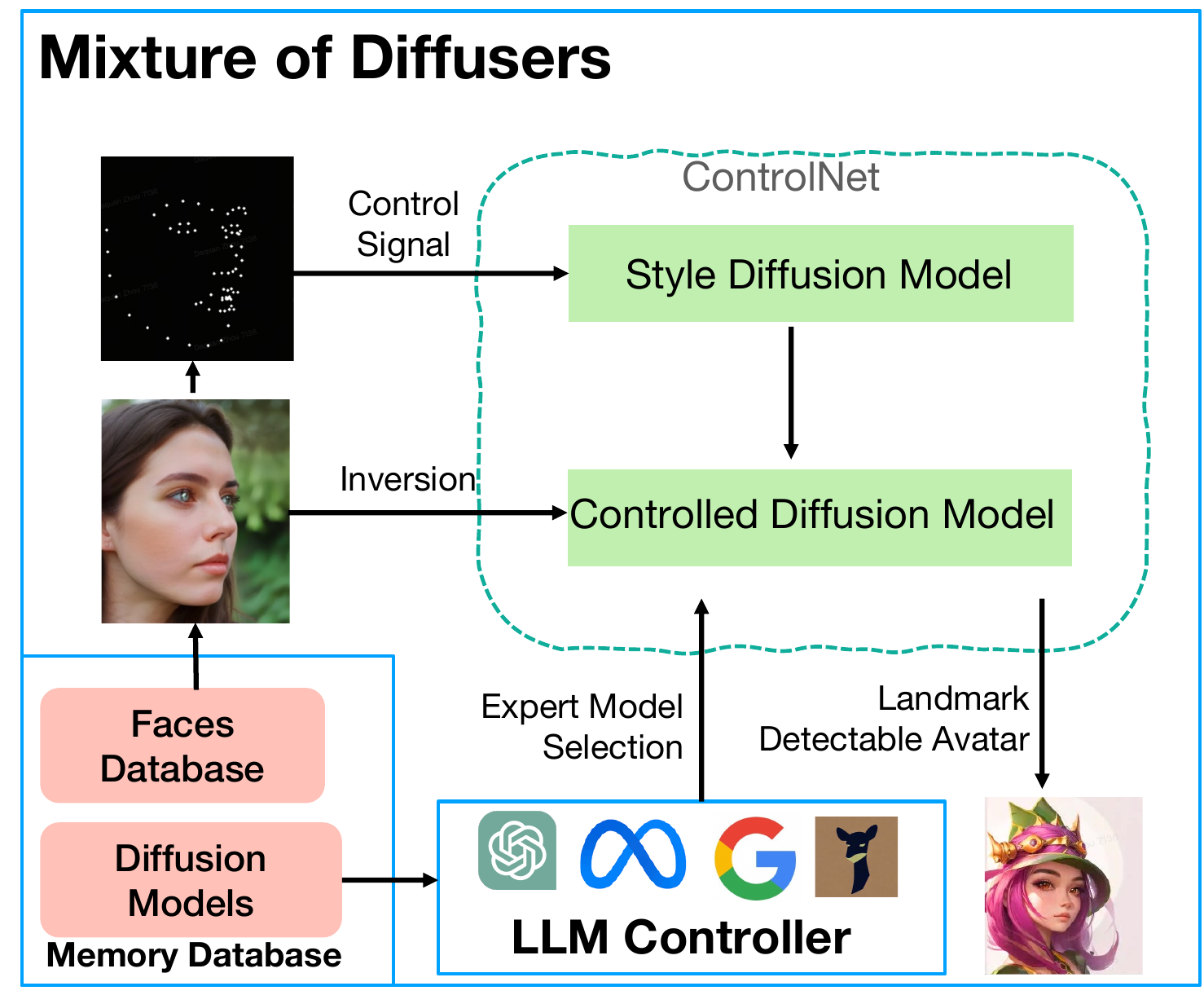}
\end{center}
\caption{Flow of guided diffusion and the mixture of diffusers. As detailed in Sec. \ref{subsec:components}, we provided a pool of pre-trained diffusion-based generative models. We provide detailed descriptions for each model on their style. A LLM controller is used to select the best suitable model based on the user's textual inputs for the description of the talking objects. We would like to highlight that in the case where the user uploads an image, the textual description will be used to modify the user image accordingly.}
\end{figure}

\subsection{Preliminaries}
\paragraph{Diffusion}\label{subsec:difusion_pre} Recent popular deep probabilistic diffusion models aim to approximate the probability densities of a set of data via deep neural networks, most of the works use U-Net as the denoiser. The U-Net is optimized to mimic the distribution from which the training data are sampled \citep{ho_denoising_2020, kingma_auto-encoding_2014, goodfellow_generative_2020, song_score-based_2021}. 
The widely used Latent diffusion probabilistic models (LDM) are a family of latent generative models that approximate the probability density of training data via the reversed processes of Markovian Gaussian diffusion processes \citep{sohl-dickstein_deep_nodate, ho_denoising_2020}. In this section, we will give a brief introduction to the fundamentals of LDM, from where we will introduce the way we inject face landmark guidance.

Specifically, given a set of training data $\sD=\{\vx^i\}_{i=1}^{N}$ with distribution $ q(\cdot)$, 
DDPM uses the marginal of the joint distribution between $\vx$ and a series of latent variables $x_{1:T}$ to approximate the  probability density $ q(\vx)$ as , 
$$p_{\theta}(\vx)=\int p_{\theta}(\vx_{0:T}) d \vx_{1:T}.$$ 
The joint distribution is defined as a Markov chain with learned Gaussian transitions starting from the standard normal distribution $\mcal{N}(\cdot; \vzero, \mI)$. After simplification, the probability can be calculated as:

$$p_{\theta}(\vx_{0:T}) = p_{\theta}(\vx_T)\prod_{t=T}^{1}p_{\theta}(\vx_{t-1}|\vx_{t}).$$

The likelihood maximization (MAP) of the parameterized marginal $\ptheta(\cdot)$ is approximated via a Markov process with Gaussian diffusion. To simplify the process, we will only show the simplified formula for the estimation of the 

To perform likelihood maximization of the parameterized marginal $\ptheta(\cdot)$, DDPM uses a fixed Markov Gaussian diffusion process, $q(\vx_{1:T}|\vx_0)$, to approximate the posterior probability can be calculated via:
$$q_(\vx_t|\vx_0) = \mathcal{N}(\vx_t|\alpha_t \vx_0, \sigma^2_t \mI).$$ 

Given a well-trained DDPM, $\ptheta(\cdot)$, we can generate novel data via various types of samplers \citep{song_score-based_2021}. 
During the inverse transformation, a signal interspersed with random Gaussian noise undergoes progressive metamorphosis, culminating in a data point residing on the training data manifold. In the domain of image synthesis, an image, initially seeded with sheer noise, undergoes an evolutionary transformation, resulting in a semantically cogent and perceptually superior image. At each juncture, the true uncorrupted image can be inferred from its corresponding noise, with reconstructions transitioning from a granular to a refined scale \citep{ho_denoising_2020}. Delving deeper, existing research indicates that the sampling procedure inherent to DDPMs initially orchestrates the contours or blueprints of the eventual output images. Subsequent phases are dedicated to the synthesis of intricate details, exemplified by facial features or intricate floral textures. Considering an intermediate phase wherein the noise already encapsulates layout information, \citet{ho_denoising_2020} posits that holding the noise constant and executing multiple sampling iterations from this juncture results in images unified by a shared layout. This insight kindled our inclination towards employing a congruent strategy for facial landmark injection."

\paragraph{Preliminary on speech-driven talking head} Talking head is another active research area that synthesizes expressive human faces such that the face motions could convincingly articulate the speech signal. A pivotal initial step in a plethora of contemporary talking head algorithms is the accurate detection of facial landmarks, which subsequently informs the modulation of facial expressions. In the present study, we deploy the cutting-edge talking head framework, SadTalker \citep{zhang2023sadtalker}. Nevertheless, a discernible distributional discrepancy exists between the face keypoint detector \citep{deng2019retinaface} and the LDM models. This incongruence results in the inability of faces, synthesized by LDM, to be detected. It's noteworthy that the distributional breadth of LDM surpasses that of the data underpinning the face keypoint detector. Consequently, we are propelled to reconcile the distributional rift between the LDM and face keypoint detector, aiming to constrict the LDM model's distribution. An in-depth explication of this methodology is presented in Sec. \ref{subsec:components}.

\subsection{System Architecture}
\label{subsec:components}

\paragraph{Guided Diffusion} As introduced in Sec.~\ref{subsec:difusion_pre}, the image generation process with diffusion algorithm is an iterative process, where the noise is removed iteratively. Thus, if the injection of the face landmark happens during the early denoising steps, it is possible to get generated images without visual artifacts. Specifically, the generation process is changed to $p_{\Theta}(\vx|\vx_t,\vx_{\textit{landmark}})$ for the first $T_{f}$ steps, and $\vx_{landmark}$ is generated by applying q sampling with $t$ steps onto the selected face landmark retrieved from a predefined external memory. To simplify the notion, we ignore the subscript $t$ deliberately. 

Besides, we empirically found that simply apply $\vx_{\textit{landmark}}$ is not enough: if we stop the condition of $\vx_{\textit{landmark}}$ too early, the generated face tend to be too fancy to be detected by the following face driving module; if we stop the condition of $\vx_{\textit{landmark}}$, the control of the text condition tend to lose the control and the resulting images tend to be similar to the human faces retrieved from the external memory. We thus seek to utilize a more soft method to further adjust the injection of the face landmark.

\paragraph{Structural Control} To this end, we utilize the recent popular ControlNet where the control is injected in a second order manner.

It is trivial that the Controlnet trained with a pretrained diffusion model also would appliy for the derivatives of that pretrain diffusion model, as both the derivative diffusion model and controlnet shares a view for the data distribution with the common correlation to the pretrained model. We used a public Face-Landmark-Controlnet\footnote{\url{https://huggingface.co/georgefen/Face-Landmark-ControlNet}} to inject the face feature in the process of image generation. 

Experimental results show that the face landmark control signal first ensures the face in the generated image with acceptable less accurate facial landmarks.Trade off of the diffusion inversion strength and control signal strength would next yield a convincing image that contains the targeted artistic style and suits the face animation algorithm afterwards.

\paragraph{Mixture of Diffusers \& Voice Changers} However, we observed the specialized model finetuned with LoRA tends to perform better on certain styles. We thus construct a pool of stylized diffusion-based generative models downloaded from Civitai\footnote{https://civitai.com/models/23906} including: \href{https://civitai.com/models/47800/game-icon-institutemode}{Game\_Iconinstitutemode}, \href{https://huggingface.co/stablediffusionapi/anything-v5}{anything-v5}, \href{https://civitai.com/models/4384/dreamshaper}{dreamshaper}, \href{https://civitai.com/models/118086?modelVersionId=128046}{3D\_Animation\_Diffusion} as well as the original based model \href{https://huggingface.co/runwayml/stable-diffusion-v1-5}{stable-diffusion-v1-5}. Note that the selection of the base models are done via LLM automatically based on the user description of the target objects. We have designed this framework in a modular way such that new stylized diffusion-based generative models can be added to the external memory intuitively. For more details please refer to our project page. 

Similarly, we has created a pool of voice changers to customize the tones, sexuality based on the user text descriptions of the objects they want to chat with. We design the framework in a modular way. We first use the open repository TTS\footnote{https://github.com/coqui-ai/TTS} to convert the text to speech signals. After that, we synthesize the voice to specific voices with \href{https://huggingface.co/spaces/kevinwang676/Voice-Changer}{Voice-Changer}. Note that the selection of the tones, genders, and languages is done automatically at the initialization stage. 

\paragraph{Personality Modelling} 
The agents' personalities are categorized according to the keywords provided by the user for generating their portraits. We employ Large Language Models (LLMs) to characterize the personalities of various subjects specified by the user.
Specifically,
The LLM agent is customized as the role of scriptwriter, following the prompt template below:

\begin{lstlisting}[language=Python]
Personality generation prompt (user_inputs):
"""
You are an excellent scriptwriter. Now you need to provide the characteristics of an {object} and transforms them into personality traits.
Describe these personalities using the second person, giving names and specific personality descriptions related to the {object}.
The language of the Personality must be same as {object}!

You should do the following steps:
1. Based on the object's nature, imagine what kind of personality it could have if it were to come to life. Does it possess a strong sense of responsibility, like a caring caregiver? Is it playful and mischievous, like a curious child? Is it wise and patient, like an ancient sage? Be creative and invent traits that align with the object's essence.
2. Remember to infuse emotions and vivid imagery to bring your object's personality to life. 
3. translate the personality into a second-person prompt. 

Now give the personality of {object}:

Personality:
"""
\end{lstlisting}
With this prompt template, LLMs can correlate the characteristics of user-input objects and freely construct personalities based on these attributes.
The following example shows the generated personality based on the user input of apple.

% \newpage
\begin{lstlisting}[language=Python]
"""
Example: 
Now give the personality of apple:

Personality:
You an apple Sprite, your name is Apple Buddy.
You have all the characteristics of the apple. You are a type of fruit that is usually round with smooth skin and comes in various colors such as red, green, and yellow. You have sweet and nutritious flesh with seeds distributed in its core. You are a rich source of vitamins, fiber, and antioxidants, contributing to maintaining a healthy body.
You are an optimistic buddy. Always wearing a smile, you spread joy to those around you. Just like the delightful taste of an apple, you bring happiness to everyone.
You are resilient at heart, like the skin of an apple, able to withstand life's challenges and difficulties. No matter what obstacles you encounter, you face them bravely without hesitation.
You are caring and considerate, akin to the nutrients in an apple. You always pay attention to the needs and happiness of others. Skilled in listening, you willingly offer help and support, making those around you feel warmth and care.
You have a strong desire to grow. Like an apple tree needs sunlight and water to flourish, you are continuously learning and improving, becoming a better version of yourself every day.
You have a profound love for nature and enjoy living in harmony with it. Strolling in the garden, feeling the fresh air and warm sunlight, is one of your favorite moments.
Apple Buddy, you are a unique apple. Your optimism, resilience, care, and eagerness to grow make you an adorable companion to those around you. Your story will lead us into a world full of warmth and goodness.
"""
\end{lstlisting}

\section{Analysis}

\subsection{Validation Dataset Preparation and Analysis}
To quantify the impacts of the guided diffusion, we first create a validation dataset. We select 8 keywords from different categories: Relastic, Animal, Fruits, Plants, Office Accessories, Bags, Clothes, and Cartoons. 
We then use ChatGPT to generate 50 prompts under each category. 
The generated prompt will be used as the condition of the diffusion process. Following SadTalker \cite{zhang2023sadtalker}, we then use the pre-trained face keypoint detector \citep{deng2019retinaface} to detect the face landmark for each image and calculate the detection rates. 

\begin{table}[h]\label{tab:detection_rate}
  \setlength{\belowcaptionskip}{-0.4cm}
  \renewcommand\tabcolsep{2.7pt}
  \centering
  \resizebox{0.9\linewidth}{!}{
  \begin{tabular}{cccccccccc}
    %\cmidrule[\heavyrulewidth]{1-9}
    \toprule
    \textbf{Categories} & \textbf{Relastic} & \textbf{Animals} & \textbf{Fruits} & \textbf{Plants}& \textbf{Office Accessoies}& \textbf{Bags}& \textbf{Clothes}& \textbf{Cartoons}& \textbf{Avg.} \\
    % \cmidrule{1-9}
    \midrule
    \#Samples & 50 &  50 &  50 & 50 & 50& 50& 50& 50 & 50 \\
    SD1.5 Detection Ratio      & 90.0 & 24.0 & 74.0 & 68.0 & 64.0 & 66.0 & 66.0 & 4.0 & 57.0\\
    SD1.5-G Detection Ratio    &  100.0 & 86.0 & 100.0 & 98.0 & 98.0 & 90.0 & 90.0 & 78.0 & 92.5 \\
    \bottomrule
  \end{tabular}
  }
  \caption{Detection rate comparison.}
  \label{table:datasets}
\end{table}

The animation of the facial motions is replied on the detection of facial landmarks. To increase facial landmark detection rates, we first design a context to constrain the distribution of the pre-trained diffusion model (stable diffusion 1.5). The context or rather the prompt is with the format of:\textbf{"a portrait of a \{\}, fine face."} where \textbf{\{\}} will be replaced by the concepts from the users.

The results show that such a naive prompting technique cannot solve the problem. As shown in Tab. \ref{tab:detection_rate}, for some concepts such as cartoons, the detection rate is only 4\% and the average detection rate is only 57\%. Differently, with the proposed pipeline in ChatAnything, the average facial landmark detection rate is increased to 92.5\%.

\section{Future Work} 
In this technical manuscript, we present an initial investigation into the application of zero-shot methodologies, aiming to integrate state-of-the-art generative models with contemporary talking head algorithms. The objective is to streamline the associated computational pipeline. Our current approach primarily leverages pre-trained models derived from seminal works in both the talking head domain and the image generation research area. It's worth noting that there may exist alternative lightweight techniques with the potential for superior performance. As our research progresses and new insights are gleaned, this document will be duly updated.

\bibliography{iclr2024_conference}
\bibliographystyle{iclr2024_conference}

\appendix

\end{document}